\documentclass{article}

\PassOptionsToPackage{numbers, compress}{natbib}
\usepackage[preprint]{Styles/neurips_2024}

\usepackage[utf8]{inputenc} % allow utf-8 input
\usepackage[T1]{fontenc}    % use 8-bit T1 fonts
\usepackage{hyperref}       % hyperlinks
\usepackage{url}            % simple URL typesetting
\usepackage{booktabs}       % professional-quality tables
\usepackage{amsfonts}       % blackboard math symbols
\usepackage{nicefrac}       % compact symbols for 1/2, etc.
\usepackage{microtype}      % microtypography
\usepackage{xcolor}         % colors

\usepackage{multirow}
\usepackage{graphicx}
\usepackage{amsmath}
\usepackage{bm}
\usepackage{wrapfig}
\usepackage{pgffor}
\usepackage{pgfmath} % For arithmetic operations

\newcommand\ours[0]{LucidPPN}

\title{\ours: Unambiguous Prototypical Parts Network for User-centric Interpretable Computer Vision}

\author{%
  Mateusz Pach\\
  Faculty of Mathematics \\ and Computer Science\\
  Jagiellonian University\\
  Kraków, Poland \\
  \texttt{mateusz.pach@student.uj.edu.pl} \\
  \And
  Dawid Rymarczyk\\
  Faculty of Mathematics \\ and Computer Science\\
  Jagiellonian University\\
  Kraków, Poland \\
  Ardigen SA \\
  \texttt{dawid.rymarczyk@uj.edu.pl} \\
  \And
  Koryna Lewandowska\\
  Department of Cognitive Neuroscience \\ and Neuroergonomics \\
  Institute of Applied Psychology\\
  Jagiellonian University\\
  Kraków, Poland \\
  \texttt{koryna.lewandowska@uj.edu.pl} \\
  \And
  Jacek Tabor\\
  Faculty of Mathematics \\ and Computer Science\\
  Jagiellonian University\\
  Kraków, Poland \\
  \texttt{jacek.tabor@uj.edu.pl} \\
  \And
  Bartosz Zieliński\\
  Faculty of Mathematics and Computer Science\\
  Jagiellonian University\\
  Kraków, Poland \\
  NCBR IDEAS \\
  \texttt{bartosz.zielinski@uj.edu.pl} \\
}

\begin{document}

\maketitle

\begin{abstract}
Prototypical parts networks combine the power of deep learning with the explainability of case-based reasoning to make accurate, interpretable decisions. They follow the this looks like that reasoning, representing each prototypical part with patches from training images. However, a single image patch comprises multiple visual features, such as color, shape, and texture, making it difficult for users to identify which feature is important to the model.
To reduce this ambiguity, we introduce the Lucid Prototypical Parts Network (LucidPPN), a novel prototypical parts network that separates color prototypes from other visual features. Our method employs two reasoning branches: one for non-color visual features, processing grayscale images, and another focusing solely on color information. This separation allows us to clarify whether the model's decisions are based on color, shape, or texture. Additionally, LucidPPN identifies prototypical parts corresponding to semantic parts of classified objects, making comparisons between data classes more intuitive, e.g., when two bird species might differ primarily in belly color.
Our experiments demonstrate that the two branches are complementary and together achieve results comparable to baseline methods. More importantly, LucidPPN generates less ambiguous prototypical parts, enhancing user understanding.
\end{abstract}
\section{Introduction}

Increased adoption of deep neural networks across critical fields, such as healthcare~\cite{rymarczyk2022protomil}, facial recognition~\cite{mollahosseini2016going}, and autonomous driving~\cite{wu2017squeezedet}, shows the need to develop models in which decisions are interpretable, ensuring accountability and transparency in decision-making processes~\cite{rudin2019stop,rudin2022interpretable}. One promising approach is based on prototypical parts~\cite{chen2019looks,donnelly2022deformable,nauta2021neural,nauta2023pip,rymarczyk2021protopshare,rymarczyk2022interpretable,ukai2022looks,wang2021interpretable,wang2023learning}, which integrate the power of deep learning with interpretability, particularly in fine-grained image classification tasks. During training, these models learn visual concepts characteristic for each class, called Prototypical Parts (PPs). In inference, predictions are made by identifying the PPs of distinct classes within an image. This way, the user is provided with explanations in the form of ``This looks like that''.

%\todo{Bartek: Zasada do powolania: "When evaluating the interpretability of an image classification network, it is important to consider not only the model’s capability to explain its reasoning process, but also the quality of its explanations."~\cite{ma2024looks}}

The primary benefit of PP-based methods over post hoc approaches is their ability to directly incorporate explanations into the prediction process~\cite{chen2019looks}. Nevertheless, a significant challenge with these methods lies in the ambiguity of prototypical parts, which are visualized using five to ten nearest patches. Each patch encodes a range of visual features, including color, shape, texture, and contrast~\cite{ma2024looks,nauta2021looks}, making it difficult for users to identify which of them are relevant. This issue is compounded by the fact that neural networks are generally biased towards texture~\cite{geirhos2018imagenet} and color~\cite{hosseini2018assessing}, whereas humans are typically biased towards shape~\cite{de2008perceived,landau1988importance}.

\begin{figure}[]
    \vspace{-1em}
    \begin{center}
    \includegraphics[width=0.75\textwidth]{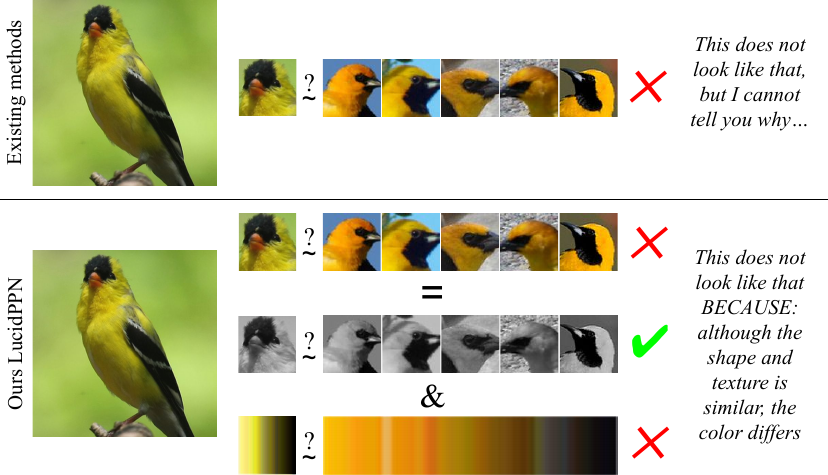}
    \end{center}
    \vspace{-1em}
    \caption{Our novel prototypical parts-based model, \ours, enables the disentangling of color information from the prototypical parts. This capability allows us to examine more closely the differences between an image patch and patches representing a prototypical part. As shown in the image, our model can visualize that the head of a bird, compared to the prototypical part of a bird's head from different classes, shows a high resemblance in shape and texture but differs in color. Such detailed analysis was not possible with previous prototypical parts-based approaches.}
    \label{fig:teaser}
    \vspace{-2em}
\end{figure}

Therefore, recent works have attempted to solve this problem using various strategies. Some papers propose to reduce the ambiguity of prototypical parts by visualizing them through a larger number of patches~\cite{ma2024looks,nauta2023pip}. However, it does not solve the problem with various visual features encoded in each patch. Other approaches tend to solve this problem by quantifying the appearance of specific visual features~\cite{nauta2021looks} or concepts~\cite{wan2024interpretable} on prototypical parts. However, they generate ambiguous statements such as ``color is important'', leading to further questions (e.g. about which color) that complicate understanding~\cite{ma2024looks,xu2023sanity}.

\begin{figure}[t]
    % \vspace{-1em}
    \begin{center}
    \includegraphics[width=0.9\textwidth]{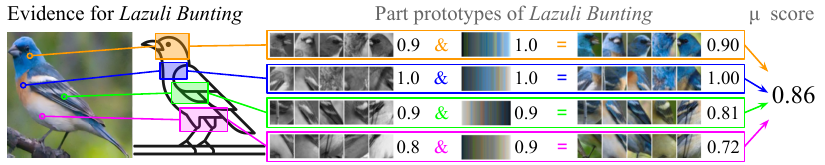}
    \end{center}
    \vspace{-1em}
    \caption{Our novel type of visualization utilizes the fact that the successive prototypes in each class of LucidPPN correspond to the same object parts. That is why we use a schematic drawing of a bird to show the location of the specific prototypical parts. Moreover, LucidPPN disentangles color features from the prototypical parts to present pairs of a simplified gray prototypical part and a corresponding color. The aggregated resemblance is obtained by multiplying the resemblance to the prototypical part and the resemblance to the corresponding color.}
    \label{fig:local}
    \vspace{-1.5em}
\end{figure}

Motivated by the challenge of decoding the significant visual attributes of prototypical parts, we introduce the Lucid Prototypical Parts Network (LucidPPN). It uniquely divides the model into two branches: the first focuses on identifying visual features of texture and shape corresponding to specific object parts (e.g. heads, tails, wings for birds), while the second is dedicated solely to color. It allows us to disentangle color features from the prototypical parts and present pairs of a simplified gray prototypical part and a corresponding color, as presented in Figure~\ref{fig:teaser}. The second advantage of LucidPPN is that the successive prototypes in each class correspond to the same object parts (e.g., the first prototypes are heads, the second prototypes are legs, etc.). Altogether, it enabled us to introduce a novel type of visualization presented in Figure~\ref{fig:local}, more intuitive and less ambiguous according to our user studies.

Extensive experiments demonstrate that LucidPPN achieves results competitive with current PPs-based models while successfully disentangling and fusing color information. Additionally, using \ours~, we can identify tasks where color information is an unimportant feature, as demonstrated on the Stanford Cars dataset~\cite{krause20133d}. Finally, a user study showed that participants, guided by \ours~explanations, more accurately identified the ground truth compared to those using PIPNet.

Our contributions can be summarized as follows:
\begin{itemize}
    \item We introduce LucidPPN, a novel architecture based on prototypical parts, which disentangles color features from the prototypical parts in the decision-making process. Consequently, thanks to \ours~we know the relevance of the color and shape with texture in the final decision process\footnote{See the discussion in paragraph Color Impact in Section 5.}.
    \item We introduce a more intuitive type of visualization incorporating the assumption about the fine-grained classification.
    \item We conduct a comprehensive examination demonstrating the applicability and limitations of LucidPPN. Specifically, we highlight scenarios where color information may not be pivotal or even can confuse the model in fine-grained image classification scenarios.
\end{itemize}

\section{Related works}

\paragraph{Ante-hoc methods for XAI.}
Self-explainable models (ante-hoc) aim to make the decision process more transparent by providing the explanation together with the prediction, and they have attracted significant attention~\cite{NEURIPS2018_3e9f0fc9,bohle2022b,brendel2018approximating}. Much of this attention has focused on enhancing the concept of prototypical parts introduced in ProtoPNet~\cite{chen2019looks} to represent the activation patterns of networks. Several extensions have been proposed, including TesNet~\cite{wang2021interpretable} and Deformable ProtoPNet~\cite{donnelly2022deformable}, which exploit orthogonality in prototype construction. ProtoPShare~\cite{rymarczyk2021protopshare}, ProtoTree~\cite{nauta2021neural}, ProtKNN~\cite{ukai2022looks}, and ProtoPool~\cite{rymarczyk2022interpretable} reduce the number of prototypes used in classification. Other methods consider hierarchical classification with prototypes~\cite{hase2019interpretable}, prototypical part transformation~\cite{li2018deep}, and knowledge distillation techniques from prototypes~\cite{keswani2022proto2proto}. Prototype-based solutions have been widely adopted in various applications such as medical imaging~\cite{afnan2021interpretable,barnett2021case,kim2021xprotonet,rymarczyk2022protomil,singh2021these}, time-series analysis~\cite{gee2019explaining}, graph classification~\cite{rymarczyk2022progrest,zhang2021protgnn}, semantic segmentation~\cite{sacha2023protoseg}, and class incremental learning~\cite{rymarczyk2023icicle}. 

However, prototypical parts still need to be improved, especially regarding the understandability and clarity of the underlying features responsible for the prediction~\cite{kim2022hive}. Issues such as spatial misalignment of prototypical parts~\cite{carmichael2024pixel,sacha2024interpretability} and imprecise visualization techniques~\cite{gautam2023looks,xu2023sanity} have been identified. There are also post-hoc explainers analyzing visual features such as color, shape, and textures~\cite{nauta2021looks}, and approaches using multiple image patches to visualize the prototypical parts~\cite{ma2024looks,nauta2023pip}. In this work, we address the ambiguity of prototypical parts by presenting a dedicated architecture, LucidPPN, that detects separate sets of prototypes for shapes with textures and another set for colors. This approach aims to enhance the interpretability and clarity of the interpretations, showcasing the rationale behind the predictions.
\vspace{-1.0em}
\paragraph{Usage of low-level vision features for image classification.}
Multiple approaches to extracting features based on texture~\cite{armi2019texture,haralick1973textural}, shape~\cite{khan2012modulating,mingqiang2008survey,rautkorpi2004novel}, and color~\cite{chen2010adaptive,kobayashi2009color} have been proposed before the deep learning era. These features are handcrafted based on cognitive science knowledge about human perception~\cite{fan2017image}. Recent studies have explored how deep learning perception models differ from human perception, revealing that neural networks can be biased toward texture~\cite{geirhos2018imagenet} and color~\cite{hosseini2018assessing}, while humans are biased toward shape~\cite{de2008perceived,landau1988importance}. These techniques have been applied in the Explainable AI (XAI) field to develop post-hoc explainers for better understanding self-supervised learned models~\cite{basaj2021explaining,laina2022measuring,rymarczyk2022comparison,zielinski2021pushes}, and prototypical parts~\cite{nauta2021looks}. We aim to build an ante-hoc model based on prototypical parts to separately process two types of visual features (texture with shape and color), and this way decreases the ambiguity of explanation.
%Our work relates to attempts to separately process color and other visual features in continual learning~\cite{berga2020disentanglement}.

\section{Method}

Our method section begins by clearly formulating the problem that this work addresses. Next, we provide a detailed overview of PDiscoNet~\citep{van2023pdisconet}, a crucial component utilized in our approach. Finally, we describe \ours, ensuring that each step of our methodology is thoroughly explained.

\vspace{-1em}
\paragraph{Problem formulation}
Our objective is to train a fine-grained classification model based on prototypical parts, which accurately predicts one of $M$ subtly differentiating classes. We use $N$ image-label pairs $\{(x_0, y_0), \ldots, (x_N, y_N)\}\subset I\times \{1,\ldots, M\}$ as a training set to obtain the model returning highly accurate predictions and lucid explanations. For this, we separate color from other visual features at the input and process them through two network branches with separate sets of PPs.

\vspace{-1em}
\paragraph{PDiscoNet}
PDiscoNet~\cite{van2023pdisconet} generates segmentation masks of object parts, used in training of \ours to align $K$ successive prototypical parts of each class with $K$ successive object parts. We decided to use it instead of human annotators because it is more efficient and cost-effective. However, it can be replaced with any method of object part segmentation due to the modularity of our approach.

PDiscoNet model $f_{Disco}$ utilizes a convolutional neural network (CNN) to generate a feature map $Z_{Disco}=(z_{ij})\in (\mathbb{R}^{D_{Disco}})^{H_{Disco} \times W_{Disco}}$ from a given image $x$. Each of $H_{Disco} \times W_{Disco}$ vectors from such feature map is then compared to trainable vectors $q^{k}\in \mathbb{R}^{D_{Disco}}$ representing $K$ object parts and background, using similarity based on Euclidean distance
\vspace{-0.5em}
\begin{equation}
t_{ij}^{k} = \frac{\exp(||-z_{ij}-q_{k}||^2)}{\sum_{k'=1}^{K+1} \exp(||-z_{ij}-q_{k'}||^2)},
\label{eq:attn_map}
\end{equation}
for $i = 1,\ldots,W_{Disco}$ and $j=1,\ldots,H_{Disco}$, and $k\in\{1,\ldots,K+1\}$. This way, we obtain an attention map $T^k=(t^k_{ij})\in\mathbb{R}^{H_{Disco} \times W_{Disco}}$ for each object part and background.
Such attention maps are multiplied by feature map $Z_{Disco}$ and averaged to obtain one vector per object part. Those vectors are passed to the classification part of PDiscoNet, which involves learnable modulation vectors and a linear classifier.

A vital observation is that the maps $T^k$ continuously split the image into regions corresponding to discovered object parts thanks to a well-conceived set of loss functions added to the usual cross-entropy. They assure the distinctiveness, consistency, and presence of the semantic regions. Yet, the only annotations used in training are the class labels.

In the subsequent sections, we ignore the PDiscoNet predictions $P_{Disco}$, using only the attention maps $T^k$, which we will call \textit{segmentation masks} from now on.

\begin{figure}{}
    % \vspace{-1em}
    \begin{center}
    \includegraphics[width=0.9\textwidth]{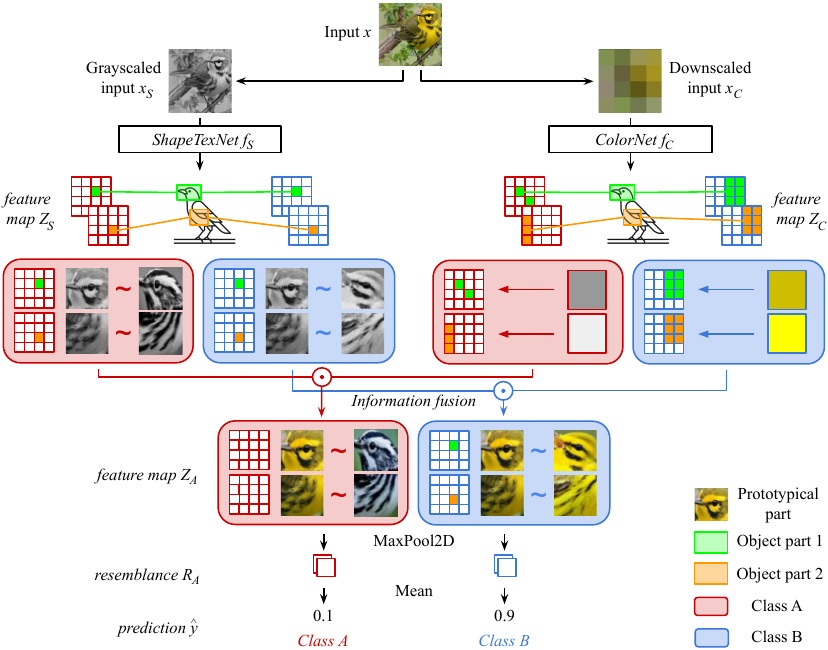}
    \end{center}
    \vspace{-1em}
    \caption{\ours~architecture consists of two branches: \textit{ColorNet} and \textit{ShapeTexNet} that encode color and shape with texture in feature maps $Z_C$ and $Z_S$, respectively. Thanks to a special type of training each channel of a feature map corresponds to similarity to a specific object part of a given class. In this image, green and orange correspond to two object parts: head and belly, and red and blue correspond to classes A and B. Therefore, each feature map consists of four channels for head of A, belly of A, head of B, and belly of B. Corresponding channels from both branches are multiplied to obtain feature map $Z_A$, which is then pooled with maximum to obtain the resemblance of prototypical parts fusion and aggregated through mean to obtain final logits.}
    \label{fig:method}
    \vspace{-1.5em}
\end{figure}

\subsection{LucidPPN} 
In this section, we first describe the architecture of our \ours. Then, we provide details on its training schema, and finally, we describe how we generate visualization of explanations.

\vspace{-1em}
\paragraph{Architecture.}
\ours~is a deep architecture, presented in Figure~\ref{fig:method}, consisting of two branches: one for revealing information about shape and texture (\textit{ShapeTexNet}), and the second dedicated to color (\textit{ColorNet}). That is why \textit{ShapeTexNet} operates on grayscaled input, while \textit{ColorNet} uses aggregated information about the color.
\vspace{-1em}
\paragraph{\textit{ShapeTexNet}.}
A grayscaled version of image $x$ is obtained by converting its channels $x=(r,g,b)$ to $x_S=(w,w,w)$, where $w=0.299r + 0.587g + 0.114b$. This formula approximates human perception of brightness~\cite{pratt2013introduction} and is a default grayscale method used in computational libraries, such as PyTorch~\cite{paszke2019pytorch}.

Grayscaled image $x_S$ is fed to a convolutional neural network backbone $f_{S_b}$. For this purpose, we adapt the ConvNeXt-tiny~\cite{liu2022convnet} without classification head and with increased stride at the two last convolutional layers to increase the resolution of the feature map, like in PIP-Net~\cite{nauta2023pip}. As an output of $f_{S_b}$, we obtain a matrix of dimension $(D\times H\times W)$, which is projected to dimension $KM\times H\times W$ using $1\times1$ convolution layer $f_{S_{cl}}$ (where $K$ is the number of object parts and $M$ is the number of classes) so that each prototype has its channel. Then, it is reshaped to the size of $K\times M\times H\times W$ on which we apply the sigmoid. As a result, we obtain \textit{ShapeTexNet feature map} defined as
\vspace{-0.4em}
\begin{equation}
Z_S = \{z_S^{km}\}_{k,m} = \sigma(f_{S_{cl}}(f_{S_{b}}(x_S))) = f_S(x_S) \in (\mathbb{R}^{H\times W})^{K\times M} 
\label{eq:ls}
\end{equation}
In consequence, we link each map $z_S^{km}$ to a unique \textit{prototypical part} of an object part $k$ for class $m$, from which we can compute \textit{ShapeTexNet resemblance} $R_S=(r_S^{km})_{k,m} \in [0,1]^{K\times M}$, where
\vspace{-0.4em}
\begin{equation}
    r_S^{km} = \text{MaxPool2D}(z_S^{km})
    \label{eq:ls}
\end{equation}
Finally, we obtain \textit{ShapeTexNet predictions} $P_S=(p_S^{m})_{m} \in [0,1]^{M}$ by taking the mean over the resemblance of all parts of a specific class
\vspace{-0.7em}
\begin{equation}
    p_S^{m} = \frac{1}{K} \sum_{k=1}^{K} r_S^{km}
    \label{eq:ls}
\end{equation}
\vspace{-1.5em}

\begin{figure}[t]
    % \vspace{-1em}
    \begin{center}
    \includegraphics[width=0.9\textwidth]{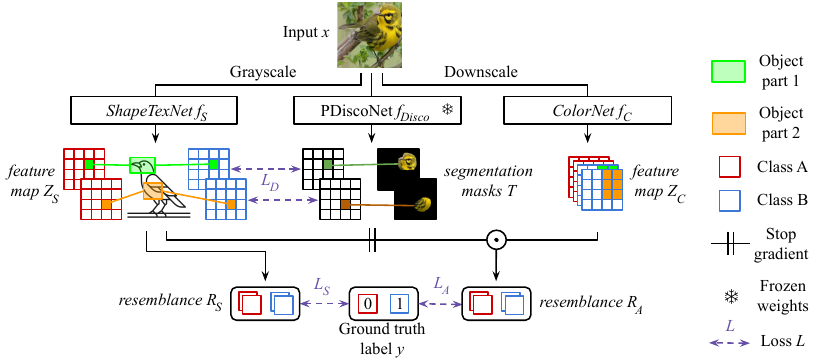}
    \end{center}
    \vspace{-1em}
    \caption{\ours~training schema. We use segmentation masks from PDiscoNet to align the activation of prototypical parts with object parts. Additionally, we enforce the \textit{ShapeTexNet} to encode as much predictive information as possible through the usage of $L_S$. Lastly, we learn how to classify images through $L_A$ which is a binary cross-entropy loss.}
    \label{fig:method}
    \vspace{-1.5em}
\end{figure}
\vspace{-1em}
\paragraph{\textit{ColorNet}.}
To obtain aggregated information about color, as an input of \textit{ColorNet}, image $x$ is downscaled to $H\times W$ resolution, marked as $x_C$. Then, $x_C$ is passed to convolutional neural network $f_C$, composed of six $1\times1$ convolutional layers with ReLU activations, except the last layer after which we apply sigmoid. This way, we process each input pixel of $x_C$ separately, taking into account only its color. As a result, we obtain \textit{ColorNet feature map} 
\vspace{-0.4em}
\begin{equation}
    Z_C= (z_C^{km})_{k,m} = f_C(x_C) \in (\mathbb{R}^{H\times W})^{K\times M}.
    \label{eq:ls}
\end{equation}
Analogously to \textit{ShapeTexNet}, each dimension in the feature map is related to a unique \textit{prototypical part} of an object part $k$ in class $m$. Hence, as before, we calculate \textit{ColorNet resemblance} $R_C=(r_C^{km})_{k,m} \in [0,1]^{K\times M}$, where
\vspace{-0.4em}
\begin{equation}
r_C^{km} = \text{MaxPool2D}(z_C^{km}).
\label{eq:ls}
\end{equation}
\vspace{-2.0em}
\vspace{-1em}
\paragraph{\textit{Information fusion and prediction}.}
To obtain \textit{aggregated feature map} $Z_A=(z_A^{km}) \in (\mathbb{R}^{H\times W})^{K\times M}$ from both branches, we multiply the \textit{ShapeTexNet feature map} with \textit{ColorNet feature map} element-wise
\vspace{-0.4em}
\begin{equation}
z_A^{km} = z_S^{km} \odot z_C^{km},
\label{eq:ls}
\end{equation}
and define \textit{aggregated resemblance} $R_A=(r_A^{km})_{k,m} \in [0,1]^{K\times M}$ as
\vspace{-0.4em}
\begin{equation}
r_A^{km} = \text{MaxPool2D}(z_A^{km}).
\label{eq:ls}
\end{equation}
% \vspace{-0.5em}
The final predictions $\hat{y} = (\hat{y}^{m})_m \in [0,1]^{M}$ for all classes are obtained by averaging $r_A^{km}$ over class-related parts
\vspace{-0.4em}
\begin{equation}
\hat{y}^{m} = \frac{1}{K} \sum_{k=1}^{K} r_A^{km}.
\label{eq:ls}
\end{equation}
\vspace{-2.0em}

\paragraph{Training.}
As a result of LucidPPN training, we aim to achieve three primary goals: 1)~obtaining a high-accuracy model, 2)~ensuring the correspondence of prototypical parts to object parts, 3)~and disentangling color information from other visual features. To accomplish these goals, we design three loss functions: 1)~prototypical-object part correspondence loss $L_D$, 2)~loss disentangling color from shape with texture $L_S$, 3)~and classification loss $L_A$ that contribute to the final loss
\vspace{-0.4em}
\begin{equation}
L = \alpha_D L_D + \alpha_S L_S + \alpha_C L_C,
\label{eq:final_loss}
\end{equation}
where $\alpha_D,\alpha_S,\alpha_C$ are weighting factors whose values are found through hyperparameter search. The definition of each loss component is presented in the following paragraphs. Please note that we assume that PDiscoNet was already trained, and we denote $\bar{y}\in \mathbb{B}^M$ as the one-hot encoding of $y$.
\vspace{-1em}
\paragraph{\textit{Correspondence of prototypical parts to object parts.}}
To ensure that each prototypical part assigned to a given class corresponds to distinct object parts, we define the prototypical-object part correspondence loss function $L_D$. This function leverages \textit{segmentation masks} $T^k$ from PDiscoNet to align the activations of prototypical parts, represented by the \textit{ShapeTexNet feature map} $Z_S$ with the locations of object parts. Consequently, the activations from the \textit{aggregated feature map} $Z_A$ will also be aligned with these object parts. It is defined as
\vspace{-0.4em}
\begin{equation}    
L_D = \frac{1}{K}\sum_{k=1}^{K} \text{MBCE}{\left(Z_S^{ky}, T^k\right)},
\label{eq:lp}
\end{equation}
where $\text{MBCE}(u, v)$ is defined as the mean binary cross-entropy loss between two maps $u,v\in[0,1]^{H\times W}$.
\vspace{-0.4em}
\begin{equation}
\text{MBCE}(u, v) = \frac{1}{HW} \sum_{i=1}^{H}\sum_{j=1}^{W} \text{BCE}(u_{ij}, v_{ij}),
\label{eq:mbce}
\end{equation}
and $y$ is the ground truth class. We align only the maps corresponding to $y$ because the prototypical parts assigned to other classes should not be highly activated.
\vspace{-1em}
\paragraph{\textit{Disentangling color from other visual infromation}.}
To maximize the usage of information about shape and texture during the classification with prototypical parts, we maximize the accuracy of the \textit{ShapeTexNet} through the usage of binary cross-entropy as classification loss function on \textit{ShapeTexNet resemblances} values
\vspace{-0.4em}
\begin{equation}
L_S = \frac{1}{KM} \sum_{m=1}^{M}\sum_{k=1}^K \text{BCE}(r_S^{km}, \bar{y}_m).
\label{eq:ls}
\end{equation}
\vspace{-1.75em}
% \vspace{-1em}
\paragraph{\textit{Classification loss}.}
Lastly, to ensure the high accuracy of the model and to combine information from \textit{ColorNet} and \textit{ShapeTexNet}, we employ binary cross-entropy on \textit{aggregated resemblances} as our classification loss
\vspace{-0.4em}
\begin{equation}    
L_A = \frac{1}{KM} \sum_{m=1}^{M}\sum_{k=1}^K \text{BCE}(r_A^{km}, \bar{y}_m).
\label{eq:lc}
\end{equation}
\vspace{-1.5em}

\paragraph{Prediction interpretation.}
\ours~adopts the definition of prototypical parts from PIP-Net \cite{nauta2023pip}, where each prototypical part is represented by ten patches, typically activated by ten colored images from the training set. However, in \ours~, the visualization must demonstrate how each prototypical part is disentangled into color and shape with texture features. That is why we propose a method to present the disentangled visual features of a prototypical part by combining five grayscale patches, a color bar, and five colored patches. The grayscale and colored patches are selected from the training images with the highest \textit{ShapeTexNet resemblance} and \textit{aggregated resemblance}, respectively. The color bar is created by sampling RGB color values from the ten colored patches with the highest \textit{aggregated resemblance} and projecting them using t-SNE~\cite{van2008visualizing}.
Moreover, in contrast to PIP-Net, \ours~creates prototypical parts corresponding to the same object parts in all classes. Therefore, we can use the information about the specific object part location to enrich the explainability.
This visualization serves as the foundation for three types of inspections listed below.
\vspace{-1em}
\paragraph{\textit{Local (prediction) interpretation.}}
Figure~\ref{fig:local} demonstrates how \ours~classifies a specific sample $x$ into class $\hat{y}$ by examining the prototypical parts assigned to $\hat{y}$ that are disentangled into color and other visual features. The views are enhanced with pointers to regions of highest \textit{aggregated resemblance}, clearly associated with the object parts.
  \vspace{-1em}
\paragraph{\textit{Comparison explanation.}}
Users may wish to inspect and compare local explanations for multiple classes. \ours~facilitates this comparison by allowing users to compare prototypical parts of corresponding object parts, making the process intuitive, as shown in Supplementary Figure~\ref{fig:comp}.
\vspace{-1em}
\paragraph{\textit{Class (global) characteristic.}}
Disentangled prototypical parts corresponding to object parts reveal the patterns the model uses to classify a given class. This enables the identification of texture and shape features, as well as colors (see Sup. Figure~\ref{fig:flowers}), that describe a class without the need to analyze the final-layer connections, unlike other prototypical part-based approaches~\cite{chen2019looks, chen2020concept, donnelly2022deformable, ma2024looks, nauta2023pip, rymarczyk2021protopshare, rymarczyk2021interpretable}.

\section{Experimental Setup}

\paragraph{Datasets.}
We train and evaluate our model on four datasets: CUB-200-2011 (CUB) with 200 bird species~\citep{wah2011caltech}, Stanford Cars (CARS) with 196 car models~\citep{krause20133d}, Stanford Dogs (DOGS) with 120 breeds of dogs~\citep{khosla2011novel}, and Oxford 102 Flower (FLOWER) with 102 kinds of flowers~\citep{nilsback2008automated}. More details on image preprocessing are in the Supplement.
\vspace{-1em}
\paragraph{Implementation details.}
Each training is repeated 3 times. We made the code public. The size of \textit{ShapeTexNet feature map} is $768\times 28\times28$. The channel number of \textit{ColorNet}'s successive convolutional layers is $20, 50, 150, 200, 600, K\cdot M$.  The values of loss weights are found through hyperparameter search and equal $\alpha_D=1.4, \alpha_S=1.0, \alpha_A=1.0$. More details are in the Supplement. 
\vspace{-2em}
\paragraph{Metrics}
During the evaluation, we report the top-1 accuracy classification score. Additionally, we measure the quality of \textit{prototypical parts} alignment with object parts by calculating intersection-over-union (UoI). To assess the descriptiveness of \textit{ColorNet} we also propose \textit{color sparsity} defined as the percentage of $5\cdot5\cdot5=125$ uniformly sampled colors from RGB space which give resemblance $R_C$ lower than $0.5$. The fewer colors activate the \textit{ColorNet feature map}, the higher \textit{color sparsity}.
\vspace{-1em}
\paragraph{User study.}
We collect the testing examples from CUB which are correctly classified by both PIP-Net and \ours. These are joined with information about the two most probable classes per model and associated prototypical parts. 
To perform the user study, we use ClickWorker System\footnote{\url{https://www.clickworker.com/}}. Sixty workers ($30$ per method) answer the survey consisting of $10$ questions. They are asked to predict the model's decision based on the evidence for the top two output classes without the numerical scores. This approach mimics the user study presented in HIVE~\cite{kim2022hive} and is also inspired by the study performed in~\cite{ma2024looks}. More details and the survey template are in the Supplement.

\begin{table}[]
    \centering
    \small
    \caption{Comparison of accuracy of PPs-based models on $4$ datasets. \ours~achieves competitive results to all methods, and SOTA on 2 datasets. Note that, \ours~ is trained with $K=12$, and ``-`` means that the model was not evaluated on a given dataset by the authors.}
    \vspace{-1em}
    \begin{tabular}{ccccc}
    \toprule
        & CUB~\citep{wah2011caltech}   & CARS~\citep{krause20133d} & DOGS~\citep{khosla2011novel}  & FLOWER~\citep{nilsback2008automated}  \\
    \midrule
    ProtoPNet \cite{chen2019looks} & $79.2$ & $86.1$ & - & - \\
    ProtoTree \cite{nauta2021neural} & $82.2\pm0.7$ & $86.6\pm0.2$ & - & - \\
    ProtoPShare \cite{rymarczyk2021protopshare}& $74.7$ & $86.4$ & - & - \\
    ProtoPool \cite{rymarczyk2021interpretable}& $\bm{85.5\pm0.1}$ & $88.9\pm0.1$ & - & - \\
    PIP-Net \cite{nauta2023pip}& $84.3\pm0.2$ & $88.2\pm0.5$ & $\bm{80.8\pm0.4}$ & $91.8\pm0.5$ \\
    \midrule
    \ours & $81.5\pm0.4$ & $\bm{91.6\pm0.2}$ & $79.4\pm0.4$ & $\bm{95.0\pm0.3}$ \\
    \bottomrule
    \end{tabular}
    \label{tab:main_table}
    \vspace{-2.0em}
\end{table}

\section{Results}
In this section, we show the effectiveness of \ours~, the influence of the color disentanglement in the processing on the model's performance, and the results related to the interpretability of learned prototypical parts based on the user study.

\vspace{-1em}
\paragraph{Comparison to other PP-based models.} 
In Table \ref{tab:main_table} we compare the classification quality of \ours~and other PPs-based methods. We present the mean accuracy and standard deviation. We report best performing \ours~, which in the case of all datasets was trained with fixed $K=12$. Our \ours~achieves the highest accuracy for CARS and FLOWER datasets, and competitive results on CUB and DOGS. 

\begin{wraptable}{R}{0.535\textwidth}
    \centering
    \vspace{-2.em}
    \small
    \caption{Comparison of the accuracy of \textit{ShapeTexNet} to \ours~. Integrating color with other visual features proves advantageous for datasets containing objects found in nature. However, for the CARS dataset, adding color information does not enhance the model's performance. This is because color is not a significant feature when classifying vehicles, as the same car model can appear in various colors.}
    \begin{tabular}{ccccc}
    \toprule
     & CUB & CARS & DOGS  & FLOWER \\
    \midrule
    \textit{ShapeTexNet} & $80.4$ & $\bm{91.7}$ & $78.6$ & $93.6$ \\
    \ours & $\bm{81.8}$ & $\bm{91.7}$ & $\bm{78.9}$ & $\bm{95.3}$ \\		
    \bottomrule
    \end{tabular}
    \label{tab:shapetexnet_table}
    % \vspace{-3em}
\end{wraptable}

% \vspace{-2em}
\paragraph{Color impact.}
The influence of \textit{ColorNet} on \ours~predictions is shown in Table \ref{tab:shapetexnet_table}. We compare the accuracy of \textit{ShapeTexNet} with the \ours~predictions. The \textit{information fusion} enhances the results on the CUB, DOGS, and FLOWER datasets. However, it does not affect performance on the CARS dataset. This can be attributed to the characteristics of the CARS dataset, where vehicles of the same model can differ in colors, indicating that color is not critical for this task. This contrasts with the fine-grained classification of natural objects, such as birds and flowers, where color plays a significant role.
 
\begin{wraptable}{R}{0.47\textwidth}
    \centering
    \vspace{-3em}
    \small
    \caption{Robustness of the model to changes in image color. When the hue value is perturbed, the accuracy of PIP-Net drops significantly. In contrast, the accuracy drop for \ours~is only half as much, and for \textit{ShapeTexNet} none.}
    \begin{tabular}{ccc}
    \toprule
        & Original & Hue-perturbed \\
    \midrule
    PIP-Net & $83.9$ & $53.0$ \\
    \textit{ShapeTexNet} & $80.3$ & $80.3$ \\
    \ours & $81.9$ & $71.7$ \\
    \bottomrule
    \end{tabular}
    \label{tab:hue_table}
    \vspace{-0.5em}
\end{wraptable}

In Table \ref{tab:hue_table} we show the results of experiments aiming to analyze how the model is susceptible to the change of the color on the image. We report the accuracy of PIP-Net, \textit{ShapeTexNet}, and \ours~on original and hue-perturbed images from the CUB dataset. One can notice that PIP-Net is highly dependent on color information and its score drops by over $37\%$ after perturbation. At the same time \textit{ShapeTexNet} is immune to this transformation, while \ours~ loses approximately $12.5\%$ accuracy because of it. To alter hue we randomly rotate hue values in the HSV color space. After rotation, we adjust the luminosity of each pixel by proportionately scaling its RGB channels. This step is key to modifying the hue without changing the brightness perceived by humans.

% Lucidppn uzywa koloru w ostatecznosci, tylko gdy dane implikuja ze ma on znaczenie. Jesli dla danego gatunku nie ma drugiego o tym samym ksztalcie i innym kolorze, to lucidppn zalozy ze kolor nie ma znaczenia. W pewnych sytuacjach daje to lepsza generalizacje.

\begin{wraptable}{R}{0.57\textwidth}
    \centering
    \vspace{-1.5em}
    \small
    \caption{User study results indicate that users based on \ours~explanations outperform those with explanations from PIPNet to a statistically significant degree.}
    \begin{tabular}{ccccc}
    \toprule
    & Mean Acc. [\%] & \multicolumn{2}{c}{$p$-value} \\
    & $\pm$ Std. & random & PIP-Net \\
    \midrule
    PIP-Net  & $60.0\pm18.1$ & $0.002$ & $-$ \\
    \ours    & $\bm{67.9\pm16.9}$ & $\bm{2.13\cdot10^{-6}}$ & $\bm{0.044}$ \\
    \bottomrule
    \end{tabular}
    \label{tab:user_table}
    \vspace{-0.5em}
\end{wraptable}

\vspace{-0.5em}
\paragraph{User study.} Statistics from the user study assessing the lucidity of explanations generated by \ours~and PIP-Net are in Table \ref{tab:user_table}. We report the mean user accuracy with a standard deviation and $p$-values. Users basing their responses on \ours~explanations score significantly better than both PIP-Net and random guess baselines. We perform a one-sided $t$-test and one-sample $t$-test to compare against PIP-Net and $50\%$ accuracy, respectively.

\section{Ablation and analysis}
In this section, we analyze how the hyperparameters of the model influence its effectiveness. Specifically, we examine the impact of the number of object parts, the weights of the loss function components, and the concurrent training of both network branches on \ours~'s performance.

\begin{wrapfigure}{R}{0.5\textwidth}
    \vspace{-1em}
    \begin{center}
    \includegraphics[width=0.5\textwidth]{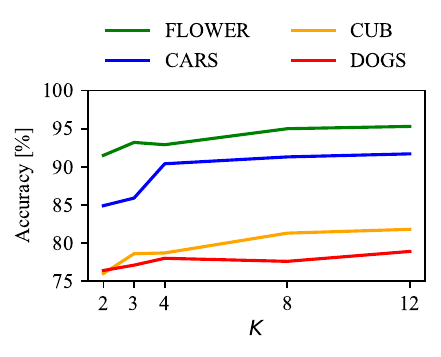}
    \end{center}
    \caption{Influence of the number of object parts $K$ on \ours~ accuracy. Increasing the number of parts improves the accuracy of the model. Note that each dataset is plotted in a unique color.}
    \label{fig:plot_k}
    \vspace{-5em}
\end{wrapfigure}

\vspace{-0.5em}
\paragraph{Number of parts.} In Figure \ref{fig:plot_k}, we show the impact of choosing different number of parts $K$. \ours~achives high results for all tested $K$, however it is noticeable that increasing $K$ improves classification. Especially on CARS, our method seems to strongly benefit from choosing $K\geq 4$. The reasonably high scores for all $K$ allow for a flexible choice between sparse explanations and higher accuracy.

\begin{wrapfigure}{L}{0.45\textwidth}
    % \vspace{3em}
    \begin{center}
    \includegraphics[width=0.45\textwidth]{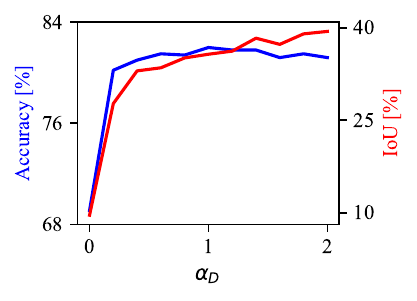}
    \end{center}
    % \vspace{-1em}
    \caption{Influence of the weight of prototypical-object part correspondence loss on accuracy and Intersection-over-Union (IoU). An increase of $\alpha_D$ improves IoU but at a certain point gradually reduces accuracy.}
    \label{fig:plot_part_weight}
    % \vspace{-2em}
\end{wrapfigure}

\vspace{-0.5em}
\paragraph{Loss weighting.} In Table \ref{fig:plot_part_weight} we investigate the impact of the loss weight $\alpha_D$, which is responsible for prototypical-object parts alignment, on training outcomes. In this analysis, the weights of the other losses are fixed at $\alpha_S=\alpha_C=1$. We evaluate the accuracy and intersection-over-union (IoU) between the highest activated \textit{ShapeTexNet feature map} and PDiscoNet's \textit{segmentation masks} for each object part. The results show that increasing $\alpha_D$ enhances the IoU, but after a certain point, it gradually reduces accuracy. Notably, omitting the loss $\alpha_D$ significantly diminishes the network's classification performance.

\begin{wrapfigure}{R}{0.45\textwidth}
    \vspace{-26em}
    \begin{center}
    \includegraphics[width=0.45\textwidth]{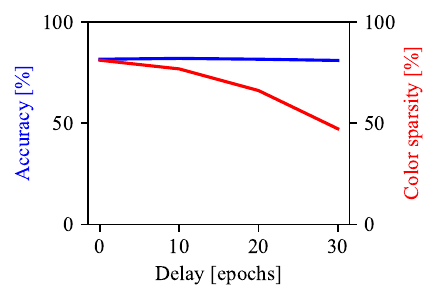}
    \end{center}
    % \vspace{-1em}
    \caption{Influence of a delay when \textit{ColorNet} starts to train on \ours~'s accuracy and color sparsity. While this delay does not negatively affect accuracy, it results in lower color sparsity. This means that the network is not concentrating on a single color when processing the PP.}
    \label{fig:plot_delay}
    % \vspace{-4em}
\end{wrapfigure}

\vspace{-0.5em}
\paragraph{Start of color network training.} It is natural to ask whether delaying the start of \textit{ColorNet} optimization could improve \ours~. In Figure~\ref{fig:plot_delay}, we report the accuracy and color sparsity after delaying the training of \textit{ColorNet}. The change in classification quality is negligible. However, we observe a drop in color sparsity, indicating that \textit{ColorNet} is less focused on relevant colors. It is important to note that despite the delay, the number of training epochs for \textit{ColorNet} remains constant for comparability.


\section{Conclusions}

In this work, we propose \ours~, an inherently interpretable model that uses prototypical parts to disentangle color from other visual features in its explanations. Our extensive results demonstrate the effectiveness of our method, and user studies confirm that our explanations are less ambiguous than those from PIPNet. In future research, we aim to further refine the model architecture to separately process shape and texture features. Additionally, we plan to explore the human perception system in greater depth to inform the design of the next generation of interpretable neural network architectures. 

\vspace{-0.5em}
\paragraph{Limitations.}
Our work faces a significant constraint: while our designed mechanism adeptly disentangles color information from input images, it cannot currently extract other crucial visual features such as texture, shape, and contrast. This highlights a broader challenge within the field: the absence of a universal mechanism capable of encompassing diverse visual attributes. Furthermore, our approach inherits limitations from other PPs-based architectures, including issues such as spatial misalignment~\cite{sacha2024interpretability} and the non-obvious interpretation of PPs~\cite{ma2024looks}. The latter could be addressed with textual descriptions of concepts discovered by PPs. 

\vspace{-0.5em}
\paragraph{Broader Impact.}
Our work advances the field of interpretability, a crucial component for trustworthy AI systems, where users have the right to understand the decisions made by these systems~\cite{kaminski2021right,tabassi2023artificial}. \ours~enhances the quality of explanations derived from PPs-based neural networks, which are among the most promising techniques for ante-hoc interpretability methods. Consequently, it can facilitate the derivation of scientific insights and the creation of better human-AI interfaces for complex, high-stakes applications.

Additionally, \ours~provides visual characteristics for PPs, which are especially beneficial in domains lacking standardized semantic textual descriptions of concepts. This is particularly useful in fields such as medicine, where it aids in analyzing radiology and histopathology images.

\section*{Acknowledgements}
The work of M. Pach, K. Lewandowska and B. Zieliński work was funded by National Centre of Science (Poland) GrantNo. 2022/47/B/ST6/03397. The work of D. Rymarczyk was funded by National Centre of Science (Poland) GrantNo. 2022/45/N/ST6/04147. The work of J. Tabor was funded by National Centre of Science (Poland) GrantNo. 2023/49/B/ST6/01137. 

We gratefully acknowledge Polish high-performance computing infrastructure PLGrid (HPC Centers: ACK Cyfronet AGH) for providing computer facilities and support within computational grant no. PLG/2023/016555. 

We are grateful to Jakub Pach for his assistance in composing images for the survey according to our developed template.

\bibliography{main}

\newpage

\section*{Supplementary Materials}
\subsection*{More details on data preprocessing}
In training, we apply transformations as follows: \texttt{Resize(size=224+8)}, \texttt{TAWideNoColor()}, \texttt{RandomHorizontalFlip()}, \texttt{RandomResizedCrop(size=(224, 224), scale=(0.95, 1.)}, where \texttt{TAWideNoColor()} is the same variation of TrivialAugment augmentation as in PIP-Net. Additionally, the image entering the \textit{ShapeTexNet} is normalized with \texttt{Normalize(mean=0.445, std=0.269)} after being converted to grayscale. At test time and when finding the prototypical parts patches, we only apply \texttt{Resize(size=224)} followed by grayscaling and normalization in case of \textit{ShapeTexNet} input. The CUB images used for training and evaluation are first cropped to the bounding boxes similarly to other PP-based methods.

We do not modify any parameters in PDiscoNet. CUB settings are used for datasets not trained in the PDiscoNet paper. For efficiency, we generated and saved the segmentation masks to avoid inferencing PDiscoNet during \ours' training.

\subsection*{More details on experimental setup}
The networks (\textit{ShapeTexNet} and \textit{ColorNet}) are optimized together in minibatches of size $64$ for $40$ epochs using AdamW~\cite{loshchilov2017decoupled} optimizer with beta values of $0.9$ and $0.999$, epsilon of $10^{-8}$, and weight decay of $0$. The learning rate of \textit{ShapeTexNet} parameters is initialized to $0.002$ and lowered to $0.0002$ after $15$ epochs. The learning rate of the \textit{ColorNet} is fixed at $0.002$. We freeze the weights of \textit{ShapeTexNet} backbone for the first $15$ epochs as a warm-up stage similar to other PPs-based approaches~\cite{chen2019looks,nauta2023pip,rymarczyk2021interpretable}.

\subsection*{More details on computing resources}
We ran our experiments on an internal cluster and a local cloud provider, a single GPU, it was either nVidia A100 40GB or nVidia H100 80GB. The node we ran the experiments on has 40GB of RAM and an 8-core CPU. The model on average trains for 3 hours.

\subsection*{More details on user study with exemplary survey}
Each worker answering a short 10-question survey was paid $1.50$ euros.
Questions between users may differ as they are randomly composed. Participants are gender-balanced and have ages from $18$ to $60$.

For PIP-Net, we randomly select samples with $K'=4,3,2,1$ in the proportion of $5:3:2:1$ based on the frequency of occurrence as PIPNet doesn't have the same number of prototypical parts assigned to data classes. The \ours~pieces of evidence for classes in the same samples always show four \textit{prototypical parts} as we use a model trained with $K=4$ here.

Example surveys for \ours~and PIP-Net are presented in Figures 14 to 26 and 27 to 39, respectively.

\subsection*{Comparison explanation example}
\begin{figure}
    \centering
    \includegraphics[width=0.6\textwidth]{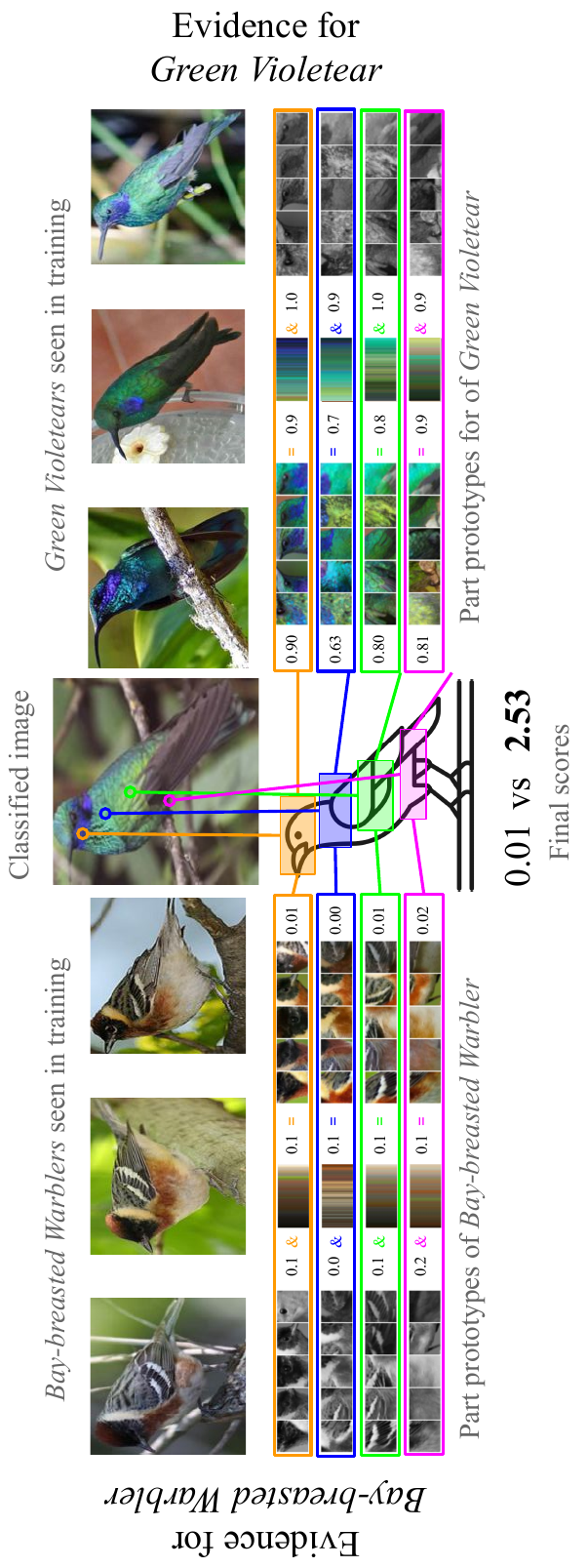}
    \caption{Comparison explanation example. Best viewed in landscape orientation.}
    \label{fig:comp}
\end{figure}

We show how our model can generate explanations as the comparison of two potential classes in Figure \ref{fig:comp}.

\subsection*{Global characteristics examples}

\begin{figure}
    % \vspace{-1em}
    \begin{center}
    \includegraphics[width=0.7\textwidth]{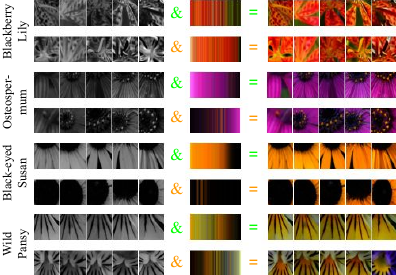}
    \end{center}
    % \vspace{-1em}
    \caption{An example showcasing global characteristics of four classes in the FLOWER dataset, using prototypical parts from \ours~trained with $K=2$. This visualization demonstrates the ability to detect differences between data classes. For instance, the \textit{osteospermum} and \textit{black-eyed susan} exhibit more variation in color, while the \textit{blackberry lilly} and \textit{wild pansay} classes differ in texture and shape.}
    \label{fig:flowers}
    % \vspace{-2em}
\end{figure}

\begin{figure}{}
    % \vspace{-1em}
    \begin{center}
    \includegraphics[width=\textwidth]{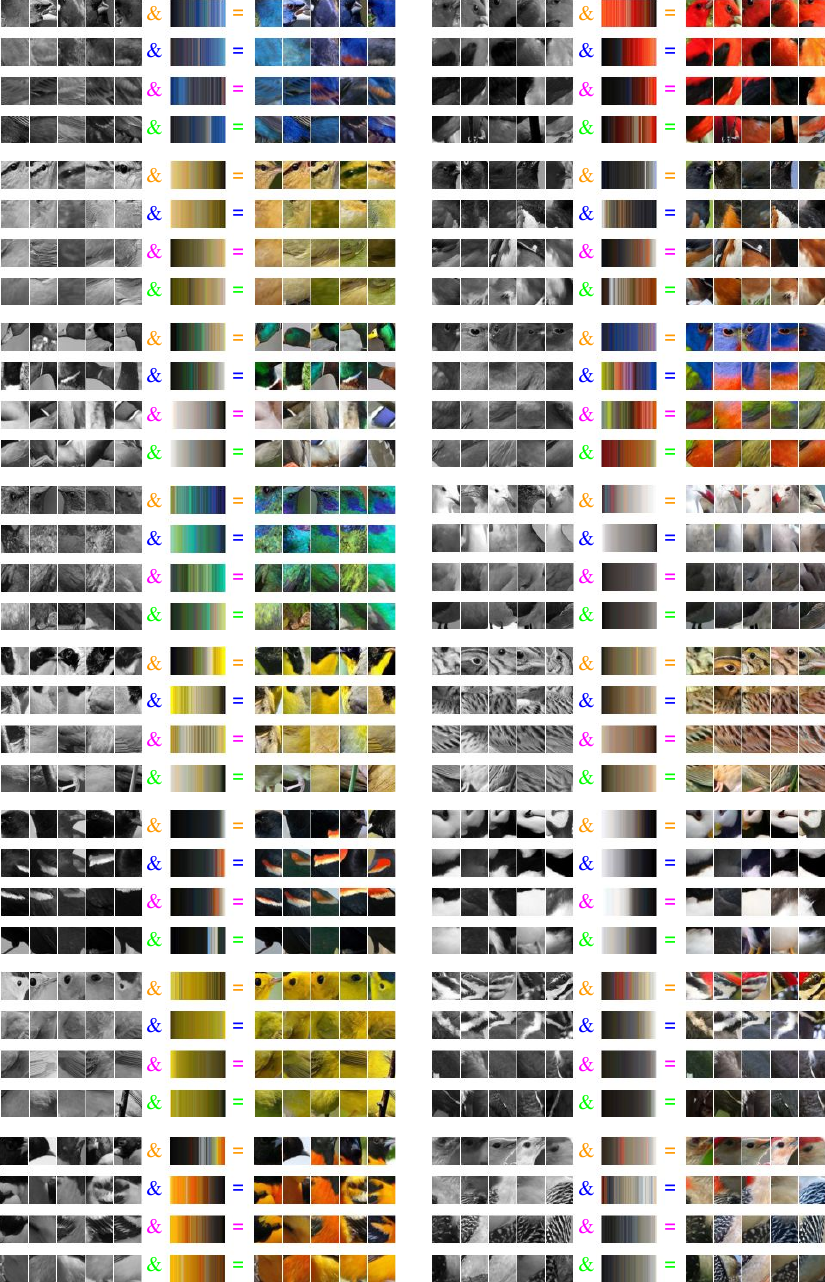}
    \end{center}
    % \vspace{-1em}
    \caption{Selected global characteristics for LucidPPN trained on CUB with $K=4$}
    \label{fig:LucidPPN-birdslong}
    % \vspace{-2em}
\end{figure}

\begin{figure}{}
    % \vspace{-1em}
    \begin{center}
    \includegraphics[width=\textwidth]{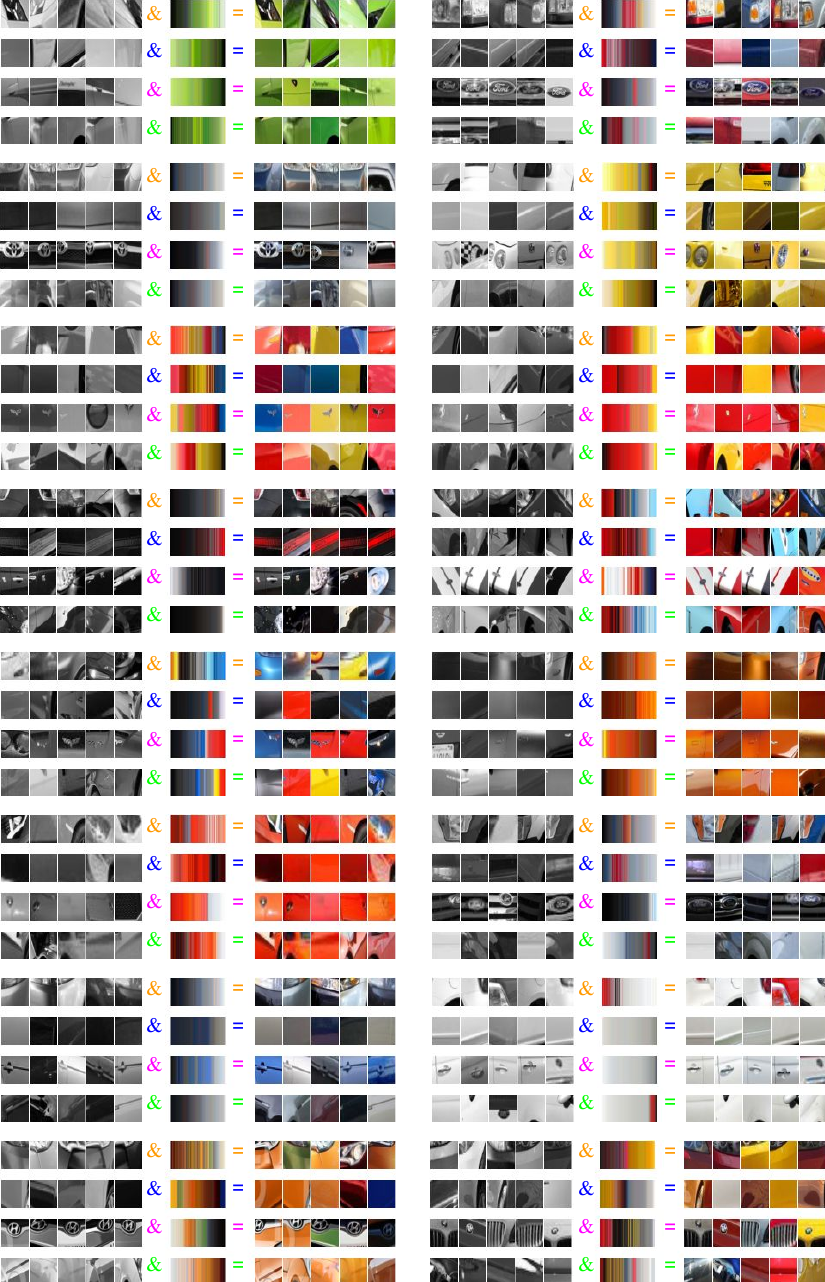}
    \end{center}
    % \vspace{-1em}
    \caption{Selected global characteristics for LucidPPN trained on CARS with $K=4$}
    \label{fig:LucidPPN-carslong}
    % \vspace{-2em}
\end{figure}

\begin{figure}{}
    % \vspace{-1em}
    \begin{center}
    \includegraphics[width=\textwidth]{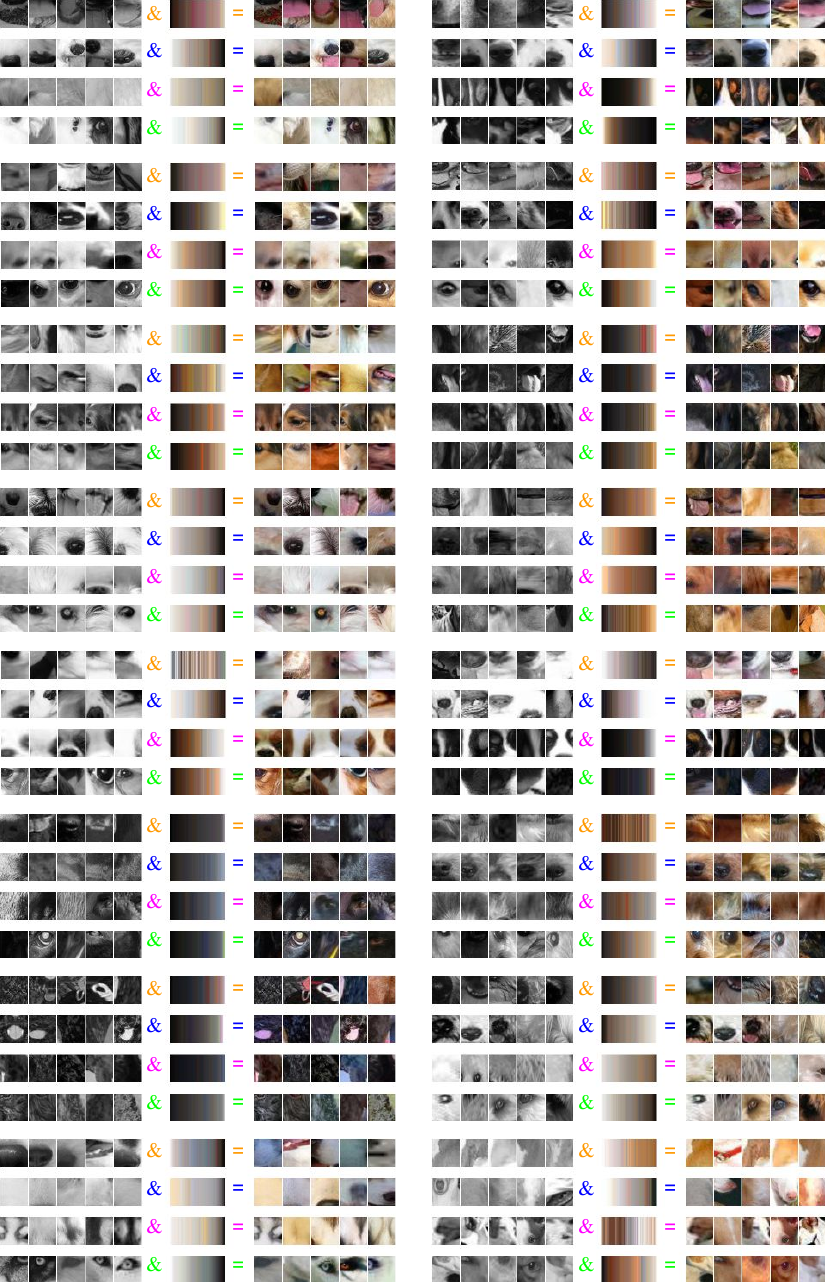}
    \end{center}
    % \vspace{-1em}
    \caption{Selected global characteristics for LucidPPN trained on DOGS with $K=4$}
    \label{fig:LucidPPN-dogslong}
    % \vspace{-2em}
\end{figure}

\begin{figure}{}
    % \vspace{-1em}
    \begin{center}
    \includegraphics[width=\textwidth]{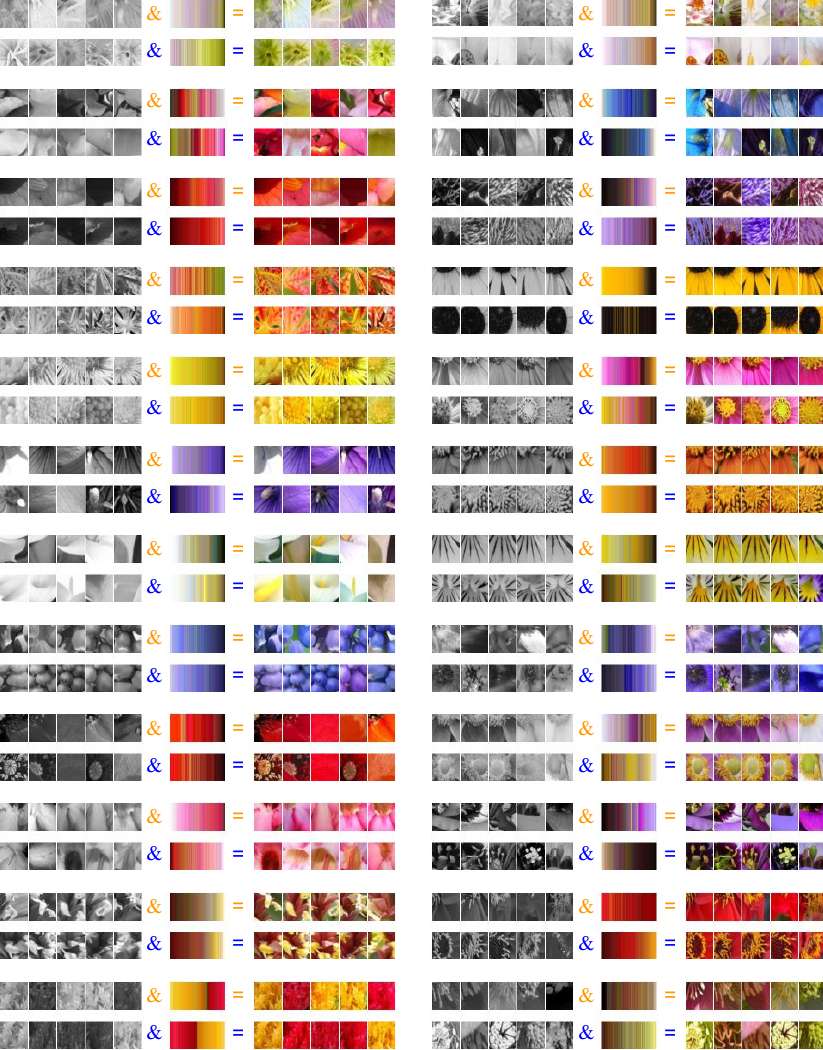}
    \end{center}
    % \vspace{-1em}
    \caption{Selected global characteristics for LucidPPN trained on FLOWER with $K=2$}
    \label{fig:LucidPPN-flowerslong}
    % \vspace{-2em}
\end{figure}

We present global characteristics for more classes and datasets in Figures \ref{fig:flowers}, \ref{fig:LucidPPN-birdslong}, \ref{fig:LucidPPN-carslong}, \ref{fig:LucidPPN-dogslong}, \ref{fig:LucidPPN-flowerslong}.

\newpage
\newpage
\newpage
\newpage

\foreach \studypagenumber in {0,...,2}{
    \begin{figure}{}
        % \vspace{-1em}
        \begin{center}
        \includegraphics[width=\textwidth]{figures/survey/lucid\studypagenumber.png}
        \end{center}
        % \vspace{-1em}
        \pgfmathparse{\studypagenumber + 4}
        \edef\result{\pgfmathresult}
        \caption{Page \pgfmathprintnumber[assume math mode=true]{\result}. of survey for \ours}
        % \label{fig:survey-lucidppn-\pgfmathprintnumber[assume math mode=true]{\result}}
        % \vspace{-2em}
    \end{figure}
}    

\foreach \studypagenumber in {0,...,2}{
    \begin{figure}{}
        % \vspace{-1em}
        \begin{center}
        \includegraphics[width=\textwidth]{figures/survey/pip\studypagenumber.png}
        \end{center}
        % \vspace{-1em}
        \pgfmathparse{\studypagenumber + 4}
        \edef\result{\pgfmathresult}
        \caption{Page \pgfmathprintnumber[assume math mode=true]{\result}. of survey for PIP-Net}
        % \label{fig:survey-pipnet-\pgfmathprintnumber[assume math mode=true]{\result}}
        % \vspace{-2em}
    \end{figure}
}

\end{document}